# Time-Scales, Meaning, and Availability of Information in a Global Brain

**Carlos Gershenson[1], Gottfried Mayer-Kress[2], Atin Das[3], Pritha Das[3], Matus Marko[4]**

[1]Centrum Leo Apostel, Vrije Universiteit Brussel
Krijgskundestraat 33 B-1160 Brussels, Belgium cgershen@vub.ac.be
[2]Dept. of Kinesiology, Penn State University, USA
[3]Dept. of Mathematics, Jadavpur University, India
[4]Faculty of Management, Comenius University, Slovakia

03/07/07

*Abstract*

We note the importance of time-scales, meaning, and availability of information for the emergence of novel information meta-structures at a global scale. We discuss previous work in this area and develop future perspectives. We focus on the transmission of scientific articles and the integration of traditional conferences with their virtual extensions on the Internet, their time-scales, and availability. We mention the Semantic Web as an effort for integrating meaningful information.

## Introduction

The time-scales involved in the transmission of *meaningful* information among humans and computers is essential for discussing the metaphor of a Global Brain [1,2]. The merits of a metaphor depend on how well it helps to unveil universal properties or patterns that are common to a large class of seemingly unrelated phenomena. In [2] one can find a description of how both biological brains as well as the Internet contain measurable parameters that control their degree of complexity - and therefore information processing capability- that the system can display. The two parameters are: 1. Size of the system as measured in the number of connected elements. 2. Characteristic time-scales that determine how fast information can spread across the network.

In complex systems theory the notion of universality expresses the observation that characteristic features of a class of systems can emerge that are independent of details of specific manifestations. Bifurcation parameters themselves are typically also unspecific in regards to the dynamics of the system itself. If several bifurcation parameters are involved in the transition (co-dimension > 1), then numerical values of the parameters, at which the transition occurs can easily be shifted by an order of magnitude or so. The number $10^{10}$ has been mentioned in [1,2] as critical number of network components characteristic for emergent properties in brains as well as in the anticipated Global Brain. Note that also the total number of neurons in some insect societies such as ant colonies has roughly the same order of magnitude and leads to some form of intelligence. In [12] it is pointed out that the number of neurons in a human brain is closer to $10^{11}$. As mentioned above, a single order of magnitude is not a big issue in discussing universal properties of complex systems. On the other hand it is clear that many neurons in the brain (part of the vegetative system etc.) are not participating in cognitive events and do not contribute to intelligence. Furthermore one could argue that with improved times-scales in information flow the critical number could also be dramatically reduced since it is balanced by faster time-scales.

We should also point out that not all of the emerging global information structures will be beneficial for the global community [13]. Today we suffer from global terrorist networks as well as daily e-mail spam, unwanted but predictable structures that emerged on the Internet just as migraines and epilepsy are the result of pathological collective firing patterns in biological brains.

In the above context "meaning" or "semantic content" is dependent on the context and is assumed to be represented by a label or marker that is attached to a term. The systematic procedure of generating these semantic labels is based on "ontologies" (see below).

The Internet has provided greatly reduced time-scales and high bandwidth (which is still increasing) for information transmission on a global scale. This can also be seen as a reduction in the time-scales required for communication, *e.g.* the time between the occurrence of a question and the actual delivery of the correct answer. Nevertheless, a general problem is that *meaningful* information, although potentially available, is difficult to find. Internet technologies have delivered several improvements in this respect over the years, but there are still many things to be done. These have not only to do with the available hardware, but also with the design of data repositories, and the interaction of humans and computers via Internet. Faster transmission of meaningful information promotes the integration of the information at a global scale. As this integration increases, the emergence of new information structures will enable us to solve problems that were not possible to solve before. If we want to achieve a higher integration, we need to improve the techniques for extracting not only meaningful but also (context dependent) "useful" information from the "noise" in the Internet. In the Global Brain metaphor, we could say that the current Internet resembles an infant brain, since it cannot yet easily differentiate between noise and useful information. The equivalent of learning and maturation will typically lead to pruning of connections so that the mature (global) brain is not continuously distracted by what appear to be irrelevant pieces of information.

Not only short time-scales and the meaningfulness of the information are required for the emergence of global information meta-structures but also its availability to the largest possible number of interested individuals and communities. It is a common case that access to important information is not free, which then leads to discrimination against economically disadvantaged populations especially in developing countries. Independent of their potential contributions they will often not be able to benefit from the integration of information, nor be able to contribute their own input. On the other hand the Internet compares most favourably with all other alternatives (paper-based or broadcast media) of global, interactive information sharing both in terms of affordability and time-scales [3].

In this paper, we review the developments of the last few years which have had an impact on the reduction of time-scales of transmission of meaningful information and its access, the actual state of affairs, and mention possible improvements for the next few years. We focus on the spreading of the information from scientific articles, the promising future of the Semantic Web for delivering meaningful information, and the general importance of time-scales as bifurcation parameter for the emergence of self-organization for instance in academic and professional conferences. Books, scientific articles and texts published on the Internet could be seen as elements of long-term memory, whereas conferences could be interpreted as cultural binding events that can lead to collective short term or "working" memory; part of which will eventually be added to the repository of the global long-term memory. In biological brains cognitive binding events are also at the basis of Hebbian learning. Under this metaphor, it is plausible to claim that the time-scales of the events taking place on the Internet should play a similarly important role than the time-scales of the events in natural brains. Of course the actual time-scales themselves will need to be adapted to the specific context:

They range from individual spikes in neurons (few milliseconds or below), gamma activities of neuronal cell assemblies (about a thousand up to tens of thousands of neurons) that participate in binding events (tens to hundreds of milliseconds), attention spans in listening to presentations (minutes to hours), collective focus of interest (days to weeks), fashions and "hot" research topics (months to years), all the way to established academic disciplines and paradigms (tens of years and longer).

## Spreading of Scientific Articles

In the early days of the Internet, when all the webpages could fit a book thinner than a phonebook, there was not a huge need of search engines or web portals, since there were relatively few things to find, and relatively few people searching for them. But with the exponential growth in information as well as users in the last few years, it is practically impossible to obtain the desired information "just surfing" or following links between related websites. Search engines and web portals have had a key role in the navigability of the Internet, since they allow the rapid delivery of meaningful and updated information to the users who request it. It is interesting to note that more advanced search engines such as Google take advantage of the "small world" effect of networked information: In a small world network it takes a surprisingly small number of steps to converge to an arbitrary target by following links between nodes ("six degrees of separation"). This phenomenon is fundamentally different from the H.G. Wells' idea of a world encyclopaedia in a traditional sense with emphasis on cataloguing information instead of linking sources. Still, most searches are still based on keywords and therefore still face the traditional drawbacks. For example, a key-word search on "complex systems" might return a site related to an office complex or to publications of chemical or psychological complexes.

Since 1999, the freely distributed Complexity Digest (http://www.comdig.org), a weekly electronic newsletter publishes edited excerpts and abstracts of recent complexity-related articles with links to the original resource on the Internet. Complexity Digest also provides webcasts of video/audio reports of conferences related to complexity. It provides quick access to recent articles and the possibility to listen to important presentations, without the need of attending remote conferences or searching in different sources. Some of the material from our webcasts has been accessed more than a thousand times in a single week (usually the week after the publication). The readers obtain relevant information related to complexity in less than a week after it becomes available. The amount of meaningful information extracted by the human editors constitutes a very small percentage compared to the amount of what would be delivered by a keyword-based search only.

Such services are becoming common in different areas; therefore reducing the time it takes a piece of information to reach the persons who might be interested in it. Nevertheless, we can see that there are still limitations to overcome.

For example, journals are making their articles available on the Internet, so that the publishing, delivery, and indexing times are skipped. But the reviewing processes in some cases can still take more than a year. Another issue is that most journals require a costly subscription, so that not all the people interested in the information can have access to it, especially in so-called developing countries. A way around this can be through the authors' initiative to share their articles by uploading them to their webpages, or to public archives. This is motivated, among others, by the Open Archives Initiative (http://www.openarchives.org). The availability also increases the impact of the article, since it becomes available to more potentially interested people. Another substantial effort is the Budapest Open Access Initiative (http://www.soros.org/openaccess) which strives to make research articles of all academic fields freely available in the Internet.

**Semantic Web**

One effort of bringing meaning to the documents on the Internet is based on a Resource Description Framework (RDF). and has been known as the "Semantic Web" [4, 5]. This meaning is directed to software agents, so that they would be able to recognize the semantic content of documents. Such a Semantic Web would increase the possibilities of interaction and would enhance the speed at which meaningful information reaches potential users, since the human editing of information will become less required. Personalized software agents could search in the Semantic Web things they recognize as relevant for their user, saving the search time for her. There are already "alarms" which make a predefined search at set intervals, and deliver the results by e-mail. These are very useful, but generally they are limited to a specific website. Searches, regular or sporadic, will deliver much better results if the search engines are able to recognize what web documents are about.

The main idea for achieving a Semantic Web involves the creation of *ontologies*. Ontologies in this context are a specification of a conceptualisation [6]. Using metadata, they help to specify inside a document what do different parts of the document mean. For example:

```
<person>Juan López</person>, who is <age>31</age> is working
as a <profession>computer engineer</profession>
```

The tags <person>, <age>, and <profession> can be "recognized" by a machine, and should not be shown to human users within a browser. Although not necessarily directly visible for human users the tags provide a basis for meaningful information extraction. This allows computer programs to organize information and retrieve it with a specified meaning. Many ontologies provide markup based on XML (Extensible Markup Language), which allows the definition (extension) of the markup. The need of explicit markup

We predict that the success of the Semantic Web will critically depend on the time scales involved in the mark-up of documents according to ontology criteria. If content creators have to spend extra efforts and time for that process then it is not very likely that the concept will find wide acceptance.

One example of the potential of a Semantic Web can be seen already with RSS (Really Simple Syndication). It is used for news syndication, making data available online for retrieval and further transmission, aggregation, or online publication. It contains only limited metadata markup, such as title, news source, description, link to the full source, and metadata about the RSS channel. This allows human users and web robots to locate much faster news, and they can use the format to sift only the ones that meet a set of relevance criteria. We are currently testing a RSS publication of the

Complexity Digest (http://www.comdig2.de/test/index.php). This has increased considerably the traffic towards the Complexity Digest.

There have been already several ontologies defined, but since natural language is changing constantly, people are beginning to study ways in which ontologies can adapt. Also, in many cases it takes a great effort to define an ontology. Because of this, different techniques are being proposed for an automated creation, or at least assistance in the creation, of ontologies (*e.g.* [7]). As mentioned above, a critical parameter for acceptance will be the time-scales associated for the users of using ontological markup or RDF-style resource description.

## Virtual Conferences

The type of interaction, which takes place during conferences is very different from the one of reading articles, basically because it is based on two-way interaction. From the metaphor of the Global Brain, we have proposed the notion of conferences as "binding events" [8], due to the similarity in the "binding together" of features (in biological brains) or different aspects or viewpoints of a complex issue (inconferences). There have been many efforts for decreasing the time-scales involved in participating in a conference ro in other words, making conference participation more effective. One of these has been the development of virtual conferences using the Internet that can enhance the information exchange function of a conference. For instance, if one potential participant can not be present during the event itself, the option of listening to a presentation in "almost real-time" can be much more effective than getting access to the printed (or CD version) of the proceedings months or even years later. Some conference try to address that problem printing proceedings ahead of time so that they are available at the conference. The effect is that often the printed material is already outdated and updated results are presented instead. To address this problem we define "almost real-time" as a time frame that is within the critical time-scale of the relevant mode of information exchange. For instance in a conference call a delay of several seconds can be perceived as not "almost realtime" whereas for conferences the critical time-scales appear to be of the order of one week (see below and [8]).

Through the Complexity Digest, we have provided the video and audio webcast of several conferences, allowing people to get an idea of the topics treated at the conference, and providing links to further material. It seems that the memory and interest of a specific presentation declines (in the average) rapidly after more than a week (at least in the area of complexity that we have data on). This also seems to agree with time-scales of "threads" in e-mail discussions where participants are only peripherally interested such as in mailing lists. More intense and longer exchanges typically happen in specialized and usually closed groups. Just like in biological brains, some events can trigger old memories that suddenly bring back details of earlier experiences. In figure 1 it looks like we have captured such a recurrence of interest, probably triggered by a complexity conference or training event. Without looking at the details of which Internet sites accessed the conference files we can only speculate about what event caused the recurrence. It is interesting that the time-scales of the recurrence is similar to that of the event itself. Furthermore it also might be a strange coincidence that the order of magnitude of accesses per week ($\sim 10^3$) is about the same as the number of neurons involved in a cell-assembly during a neuronal binding event (~$10^3$ - $10^4$). The fact that in week 0 (i.e. during the conference) we had almost 100 hits indicates that almost realtime web-casting can also be of considerable value for on-site participants. Due to parallel sessions, travel time between conference halls, saturation in attention /perception, etc, on-site participants often miss important talks especially at mega-conferences of thousands of participants and more.

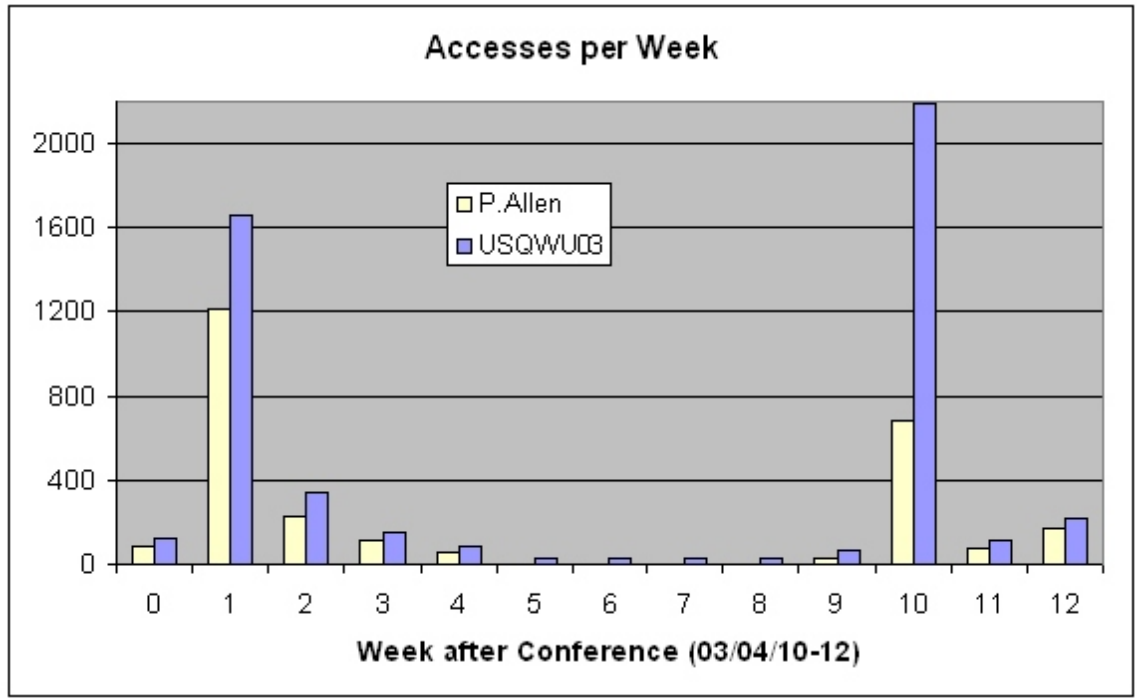

Figure 1: Weekly accesses to both audio recordings (in mp3 format) and presentation visuals (pdf or ppt format) of the conference Uncertainty and Surprise: Questions on Working with the Unexpected and Unknowable (USQWU03), The University of Texas Austin, Texas USA, 03/04/10-12, Conference files are available at: http://www.comdig2.de/Conf/USQWU2003/. This specific conference was chosen because one of the presentations (Peter Allen) holds the download record of our archive.

## Conclusions

We have mentioned actual and future ways and techniques which increase the integration of information at a global scale. The transmission of information should be *fast*, and hardware developments are achieving this. The information should be *meaningful*, and the Semantic Web is the project aiming at automatizing what humans are doing already. The information should be *available*, and projects promoting open source code and free access to scientific publications are striving for this.

The biological brain is an example of efficiency in information processing. We believe that the development of Internet systems inspired by the human brain will enhance the efficiency of this (*e.g.* [9,10]), and thus promote global integration. This integration most probably will not be made only by computers, but by the collaboration between them with humans. This is because humans and computers are better at different types of tasks. This has been already noticed and Internet systems are combining computational power with human intuition [11]. Human intuition and computer efficiency can be successfully integrated at the global level to achieve optimum results in different fields. Humans cannot perform many things computers are able to do with the same efficiency, and vice versa. But only by their mutual collaboration it will be possible to solve current and emerging global problems that are beyond our scope today.